\documentclass[sigconf,nonacm]{acmart}
\usepackage{graphicx}

\renewcommand\footnotetextcopyrightpermission[1]{}
\usepackage{booktabs}
\usepackage{amsmath}
\usepackage{enumitem}

\AtBeginDocument{%
  }

\title{Tradeoff-Aware Territory Design: A Multi-Objective Decision System for Enterprise Customer-to-Seller Allocation}

\author{Nan (Iris) Wang}
\affiliation{%
  \institution{Databricks}
  \city{San Francisco}
  \state{CA}
  \country{USA}}
\email{nan.wang@databricks.com}

\author{Mitchell Turiel}
\affiliation{%
  \institution{Databricks}
  \city{San Francisco}
  \state{CA}
  \country{USA}}
\email{mitchell.turiel@databricks.com}

\begin{abstract}
Enterprise customer journeys are shaped not only by product experiences and marketing interventions, but also by recurring human-resource allocation decisions. In B2B go-to-market (GTM) organizations, these decisions are operationalized through territory carving: the recurring process of selecting which accounts receive seller coverage, preserving valuable seller-account relationships, and allocating limited field capacity across heterogeneous opportunities. Despite its strategic importance, territory carving often remains manual and expertise-driven, making it difficult to compare account opportunities consistently, quantify the value of continuity, or reason transparently about tradeoffs among growth, balance, capacity, and operational feasibility.

We present a production decision system for tradeoff-aware customer-to-seller allocation in enterprise GTM planning. The system combines three analytical components: a multidimensional account-value layer that scores customer opportunity from multiple predictive signals; an observational continuity-valuation layer that estimates the expected cost of breaking existing account-seller relationships using a synthetic-control-style counterfactual framework; and a multi-objective optimization layer that prioritizes high-value accounts for seller coverage while balancing continuity value, opportunity balance, and operational constraints.

The framework has been deployed as a scenario-based decision-support system for enterprise territory planning. Historical backtesting shows that optimizer-generated scenarios can substantially raise assignment compliance while preserving comparable seller-level opportunity distribution. In production planning use, adopted scenarios improved both assignment compliance and opportunity balance relative to the prior planning baseline. Beyond these gains, directing scarce seller coverage toward higher-value accounts represents a meaningful incremental revenue opportunity at enterprise scale. The central contribution is less any single component than the framework it enables: one that makes these tradeoffs explicit and inspectable, turning territory carving into a governable decision surface.

\end{abstract}



\begin{document}
\maketitle
\section{Introduction}
In enterprise B2B go-to-market planning, customer-to-seller allocation is a recurring mechanism through which organizations shape the customer journey. Territory planning determines which accounts receive seller coverage, which relationships are preserved, and how scarce field capacity is allocated for the next planning period. In large enterprise GTM organizations, this process can involve many accounts, sellers, stakeholder groups, business rules, and capacity constraints. Because these decisions must ultimately result in concrete assignments, territory carving requires a decision system that can produce feasible, inspectable recommendations.

Territory carving is often distributed and expertise-driven. Business leaders make account-level decisions using domain knowledge, local context, and planning judgment. While this captures important business context, it also creates structural challenges. First, the account universe may be too large for comprehensive manual inspection. Second, account value is heterogeneous and only partially observed, making it difficult to form a shared value system. Third, account-to-seller assignment is globally coupled: assigning one account to one seller consumes capacity, changes opportunity distribution, affects relationship continuity, and constrains the remaining feasible assignment space. Fourth, scenario comparison is difficult when each alternative plan must satisfy many interacting business rules. As a result, planners have limited ability to compare tradeoffs among account value, continuity, opportunity balance, and operational feasibility.

Sales-territory design has long been studied as an alignment and resource-allocation problem, often using integer programming, heuristic optimization, or contribution-aware formulations to balance workload, opportunity, geography, and operational feasibility~\cite{zoltners1980,skiera1998,riosmercado2013}. We extend this literature by treating continuity as an empirically estimated, account-varying value rather than a fixed movement penalty, and by making heterogeneous account value directly usable in constrained optimization.

\begin{figure*}[t]
  \centering
  \includegraphics[width=\textwidth]{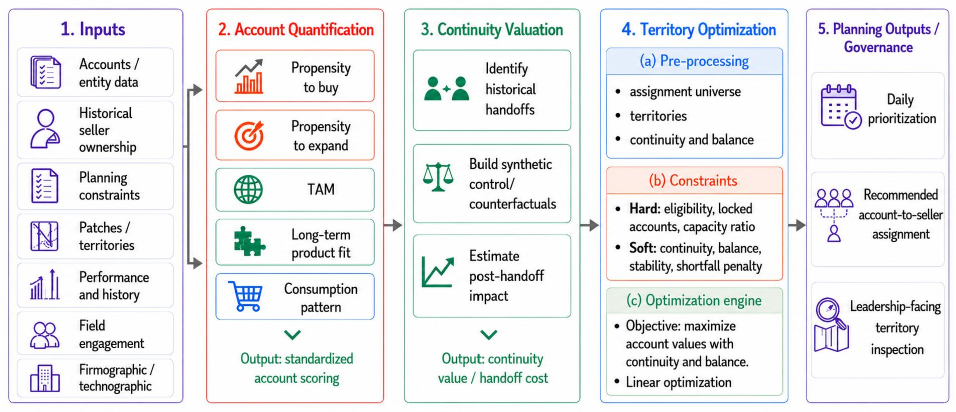}
  \caption{Overall framework for data-driven territory carving.}
  \label{fig:framework}
\end{figure*}

This paper presents a decision-system framework for tradeoff-aware territory design. The system prioritizes high-value accounts through an optimization-ready account-value measure, estimates the value of seller-account continuity using a counterfactual handoff design, and integrates these signals into a constrained optimization model for account-to-seller assignment. The resulting framework supports scenario-based planning: decision makers can start from feasible recommendations, inspect the tradeoffs they encode, and apply business judgment to exceptions. More than a particular model or solver, the contribution is a reframe: territory carving becomes an inspectable decision problem in which heterogeneous account value is represented through a common planning currency and seller continuity is treated as an empirically measured tradeoff rather than an intuitive preference.

\section{Problem Formulation}
Territory carving is not simply an account-ranking problem or a standalone assignment problem. Each account-to-seller decision has several coupled consequences: it determines whether an account receives seller coverage, consumes seller capacity, changes the distribution of opportunity across sellers, and may preserve or break an existing seller-account relationship. A useful decision system must therefore produce feasible assignments while making the underlying tradeoffs inspectable to planners.

Figure~\ref{fig:framework} summarizes the end-to-end framework and provides the roadmap for the rest of the paper. We formulate territory carving as a constrained account-to-seller assignment problem whose output is a feasible assignment plan. The optimization is driven by two upstream decision inputs: an account-value input $Q_i$, produced by the scoring layer to make heterogeneous accounts comparable for planning, and a continuity input $C_i$, interpreted as the value of preserving an incumbent seller-account relationship or, equivalently, the avoided cost of a handoff. Section~3 describes how $Q_i$ is constructed; Section~4 describes how $C_i$ is estimated; Section~5 combines these inputs in a multi-objective optimization model that generates feasible plans while navigating a tradeoff surface across opportunity capture, continuity preservation, balance, and reassignment cost~\cite{hervert2018,miettinen1999,riosmercado2013}; and Section~6 evaluates the resulting plans using plan-quality metrics rather than solver objective value alone.

Let $I$ denote the set of customer accounts and $J$ denote the set of sellers. Territory carving assigns accounts to sellers for a planning period, typically the next fiscal year. We define a binary decision variable $x_{ij}$ indicating whether account $i$ is assigned to seller $j$:

\begin{equation}
 x_{ij}=\begin{cases}
  1, & \text{if account } i \text{ is assigned to seller } j,\\
  0, & \text{otherwise.}
 \end{cases}
 \label{eq:decision-var}
\end{equation}

Each account $i \in \mathcal{I}$ is characterized by account value $Q_i$, workload weight $a_i$, eligibility conditions, and, when applicable, incumbent ownership. Each seller $j \in \mathcal{J}$ has capacity bounds and coverage eligibility.

We model territory design as a multi-objective optimization problem in which account value rewards opportunity capture, continuity value rewards relationship preservation, and balance and movement penalties discourage uneven opportunity distribution and unnecessary reassignment.

\section{Account Value Quantification}
A central challenge in territory planning is that account value is multi-dimensional and not directly observable. Current revenue may favor mature customers while understating emerging growth opportunities. Product usage may reflect realized engagement but not necessarily long-term commercial potential. Firmographic and technographic signals may indicate strategic fit, but require calibration before they can support account-level planning decisions.

To create a shared planning currency, we use a scoring framework that decomposes account value into performance, potential, and propensity. For each account $i$, we define the account value as

\begin{equation}
Q_i = f\!\left(P_i^{\text{perf}},\, P_i^{\text{pot}},\, P_i^{\text{prop}},\, P_i^{\text{fit}},\, P_i^{\text{eng}},\, \dots\right).
 \label{eq:score}
\end{equation} 

where $P^{\mathrm{perf}}_i$ captures realized business trajectory; $P^{\mathrm{pot}}_i$ captures long-term account opportunity, including total addressable market and other structural indicators; $P^{\mathrm{prop}}_i$ captures the likelihood of near-term growth or expansion within the planning horizon; $P^{\mathrm{fit}}_i$ captures strategic and firmographic fit; and $P^{\mathrm{eng}}_i$ captures account engagement, among other signals used in the composite. Predictive account scoring and customer-value modeling are widely used to prioritize scarce sales and marketing resources~\cite{gupta2006,jena2023}. In our setting, the score is designed not only for ranking accounts, but also as an optimization input for comparing heterogeneous account types.

In implementation, $f(\cdot)$ is a weighted composite of model-based scores, calibrated with planning heuristics and business priorities for the specific planning use case. The goal is not only prediction accuracy, but decision usefulness: the score must be comparable across accounts, interpretable to stakeholders, and stable enough to support territory design. By translating heterogeneous signals into a common account-value measure, the scoring framework reduces reliance on subjective prioritization and enables more consistent organizational planning decisions.

\section{Continuity Valuation Using a Synthetic-Control-Style Counterfactual}
\label{sec:continuity}
Seller-account continuity is an important but historically under-quantified dimension of territory design. Territory carving may reassign an account from an incumbent seller to a new seller because of capacity rebalancing, seller movement, or GTM redesign. While such handoffs may be operationally necessary, they can also disrupt accumulated account context and slow post-carve consumption growth. In many planning processes, continuity is treated as an intuitive preference rather than an empirically calibrated tradeoff.

We frame a seller-account handoff as a treatment event and estimate its effect on post-carve consumption using a synthetic-control-style counterfactual framework~\cite{abadie2010synthetic,abadie2021using}. Let $D_i = 1$ indicate that account $i$ experienced a seller handoff during the territory carve, and let $Y_{i,t}$ denote weekly consumption for account $i$ at time $t$. The causal estimate is the post-carve consumption impact for handed-off accounts relative to the counterfactual condition of continued seller coverage. In other words, the counterfactual represents what consumption would have been for treated accounts had the seller handoff not occurred.

Because this counterfactual is not observed, we construct an estimated no-handoff trajectory from comparable accounts that retained seller continuity. Donor accounts are selected based on pre-carve similarity in account characteristics and consumption trajectory, including spend tier, trailing consumption, and tenure. The synthetic trajectory is calibrated on the pre-carve period and then projected into the post-carve period. Let $Y^{\mathrm{obs}}_{t}$ denote observed post-carve consumption for the treated cohort, and let $\hat{Y}^{0}_{t}$ denote the estimated counterfactual consumption under continued seller coverage. The estimated handoff effect at time $t$ is
\begin{equation}
 \tau_{i,t}^{\mathrm{handoff}} = Y_{i,t} - \widehat{Y}_{i,t}^{0}.
 \label{eq:handoff-effect}
\end{equation}
Aggregating over the post-planning horizon $\mathcal{T}_{post}$ gives an account-level estimate of the handoff impact:
\begin{equation}
 \widehat{\tau}_{i}^{\mathrm{handoff}} = \frac{1}{|\mathcal{T}_{post}|}\sum_{t\in\mathcal{T}_{post}}\tau_{i,t}^{\mathrm{handoff}}.
 \label{eq:average-handoff}
\end{equation}
If handoffs reduce account performance, $\widehat{\tau}_{i}^{\mathrm{handoff}}$ is negative. We can therefore define the estimated value of preserving continuity as the avoided disruption from a handoff:
\begin{equation}
 C_i=-\widehat{\tau}_{i}^{\mathrm{handoff}}.
 \label{eq:continuity-value}
\end{equation}

\begin{figure}[t]
  \centering
  \includegraphics[width=\columnwidth]{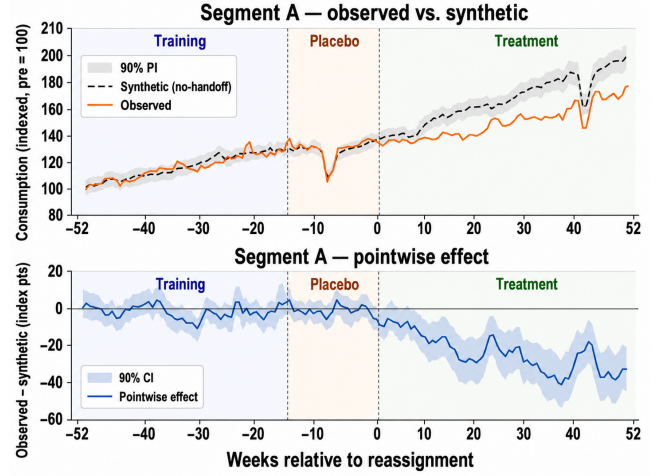}
  \caption{
  Continuity valuation using a synthetic-control-style counterfactual.
  }
  \label{fig:continuity}
\end{figure}

Figure~\ref{fig:continuity} illustrates the counterfactual design: the observed post-carve trajectory is compared against a synthetic no-handoff trajectory estimated from comparable continuity-preserved accounts. To reduce the risk of attributing residual counterfactual error or common time-period effects to seller handoff, we adjust the estimated handoff effect using placebo effects from continuity-preserved accounts.

An empirical continuity study of a recent carve shows that seller handoffs are associated with lower post-carve consumption. Across more than thousands of treated accounts, handed-off accounts consistently underperformed their synthetic no-handoff counterfactual over the following year. The estimated average treatment effect on the treated (ATT) is negative and statistically significant, and remains so after placebo adjustment, with the confidence interval excluding zero. The effect varies materially across account groups: faster-growing accounts show a roughly threefold larger estimated decline than more  established ones, and disruption is clearer in some business units and account segments than others. This heterogeneity is the central planning insight: continuity should be modeled as a differentiated  business tradeoff rather than a flat reassignment penalty, preserving seller-account relationships where disruption is expected to be costly while allowing reassignment where planning benefits outweigh the estimated continuity cost.

These estimates are observational. Seller handoffs are not randomly assigned: they arise from capacity rebalancing, seller movement, and GTM redesign, any of which may correlate with an account's trajectory. The estimate is therefore causal only under the assumption that the synthetic trajectory---built from continuity-preserved donors matched on spend tier, trailing consumption, and tenure---approximates what a handed-off account would have done absent the handoff. The placebo adjustment removes common time-period effects and residual counterfactual error estimated on untreated accounts, but it does not correct for selection into handoff on unobservables that differ between handed-off and retained accounts, nor for spillovers if reassigning one account shifts a seller's attention to others (a SUTVA violation). We therefore read the ATT as a directional, assumption-conditional estimate of handoff disruption rather than a randomized identification.

\section{Multi-Objective Territory Optimization}
\label{sec:optimization}
The optimization layer converts account value and continuity valuation into feasible account-to-seller recommendations under the constraints and tradeoffs. A key design challenge is calibration across objectives: account value, continuity, balance, and operational feasibility are not naturally expressed in the same unit. Rather than converting all terms into short-term dollars, we use account value as the optimization currency, allowing the objective to reflect both near-term commercial opportunity and long-term strategic potential. Continuity enters as an empirically estimated adjustment to the account value, so the penalty for reassignment can vary between account types.

The weighted objective can be interpreted as an inspectable scalarization of competing planning goals, a common approach in multi-objective optimization when decision makers need to compare feasible tradeoffs across objectives~\cite{miettinen1999}. Let $\mathcal{I}$ denote the set of accounts and $\mathcal{J}$ denote the set of sellers. Let $\mathcal{P}$ denote the set of feasible account-seller assignment candidates, where each candidate $p \in \mathcal{P}$ corresponds to assigning account $i(p)$ to seller $j(p)$. The decision variable $x_p$ equals 1 if assignment candidate $p$ is selected and 0 otherwise.

The purpose of the optimizer is not to claim that there is a single universally correct territory plan. Instead, the weighted objective generates inspectable points on a planning tradeoff surface. By varying the weights on account value, continuity, balance, and operational feasibility, planners can compare feasible scenarios and select an operating point that reflects current business priorities. This is especially important in territory carving because improvements along one dimension often require concessions along another: a plan may improve opportunity balance by reassigning more accounts, or preserve continuity by accepting less balance improvement. We use the following compact objective to expose the main territory-design tradeoffs:

\begin{equation}
\label{eq:objective}
\begin{aligned}
\max_{x} \quad
& \underbrace{\sum_{p \in \mathcal{P}} \alpha q_p x_p}_{\text{Account value}}
+ \underbrace{\sum_{p \in \mathcal{P}} \beta c_p x_p}_{\text{Continuity}} \\
& -\, \underbrace{\lambda \operatorname{Bal}(x)}_{\text{Opportunity balance}} 
+ \underbrace{\sum_{p \in \mathcal{P}} \gamma k_p x_p}_{\text{Intent-to-Keep}}
- \underbrace{\delta \sum_{j \in \mathcal{J}} r_j}_{\text{Shortfall}}
- \underbrace{\mu\,\Phi_{\text{op}}(x)}_{\text{operational feasibility}}.
\end{aligned}
\end{equation}

Here, $q_p = Q_{i(p)}$ is the account-value value from the scoring framework for the account in candidate assignment $p$. The continuity term $c_p$ is estimated separately from account value and translated into the same account-value currency used by the optimizer. It captures the avoided value loss from preserving seller-account continuity, with the magnitude varying by account segment, spend tier, or business unit based on the estimated handoff disruption. Thus, continuity enters the objective as an empirically calibrated, heterogeneous adjustment rather than as a flat reassignment penalty.

The balance penalty measures deviation from seller-level opportunity targets:

\begin{equation}
\label{eq:balance}
\operatorname{Bal}(x)
=
\sum_{j \in \mathcal{J}}
\left|
\sum_{\substack{p \in \mathcal{P} \\ j(p)=j}}
q_p x_p
-
T_j
\right|.
\end{equation}

The objective is optimized subject to hard GTM feasibility constraints. Each account can be assigned to at most one seller:

\begin{equation}
\label{eq:single-assignment}
\sum_{\substack{p \in \mathcal{P} \\ i(p)=i}}
x_p
\leq
1,
\quad
\forall i \in \mathcal{I}.
\end{equation}

The intent-to-keep value $k_p$ is used in production re-optimization to preserve explicit planning edits made during the current planning cycle. When field leaders adjust an assignment after reviewing an optimized scenario, the next optimization run treats that edit as planning intent rather than as noise to be reverted. This term is distinct from continuity: continuity preserves pre-existing seller-account relationships, while intent-to-keep preserves expert-reviewed assignment decisions made during the iterative planning process. The weight $\gamma$ controls how strongly the optimizer respects these edits relative to account value, continuity, balance, and other operational objectives.

Minimum workload or coverage requirements are modeled with a nonnegative shortfall variable:

\begin{equation}
\label{eq:shortfall}
\sum_{\substack{p \in \mathcal{P} \\ j(p)=j}}
a_{i(p)} x_p
+
r_j
\geq
L_j,
\quad
r_j \geq 0,
\quad
\forall j \in \mathcal{J}.
\end{equation}

Locked assignments are preserved when required by business rules. Let $\mathcal{P}^{\mathrm{lock}}$ denote the set of locked assignment candidates. Then:

\begin{equation}
\label{eq:locked}
x_p = 1,
\quad
\forall p \in \mathcal{P}^{\mathrm{lock}}.
\end{equation}

This formulation separates strategic objectives from feasibility requirements. The objective defines the tradeoff surface across account value, continuity, balance of opportunities, and coverage, while the constraints enforce the operating rules under which the territory plan must be valid. This structure makes each recommendation inspectable: a proposed territory plan can be decomposed into the account value it captures, the continuity it preserves, the balance impact it creates, and the feasibility constraints it satisfies.

\begin{table}[t]
\centering
\caption{Scenario frontier on the backtest. Imbalance reduction is measured against the deployed baseline plan; account reassignment is measured against the selected production scenario.}
\label{tab}
\begin{tabular}{p{0.36\linewidth} p{0.27\linewidth} p{0.27\linewidth}}
 \toprule
           & Balance gain & Continuity break \\
           & vs deployed plan & vs production \\
\midrule
Continuity heavier & 34\% & -46\% \\
Continuity heavy & 71\% & $-10\%$ \\
\textbf{Production (selected)} & \textbf{94\%} & \textbf{0\% (ref.)} \\
Balance heavy & 95\% & +10\% \\
Balance heavier & 95\% & +19\% \\
\bottomrule
\end{tabular}
\end{table}

Table~\ref{tab} illustrates this tradeoff surface. The selected production scenario captures most of the observed imbalance reduction relative to the deployed baseline plan. Moving to more balance-heavy scenarios adds less than one percentage point of additional imbalance reduction, but requires 10--19\% more account reassignment. Conversely, more continuity-heavy scenarios reduce reassignment but leave substantially more imbalance unresolved. We therefore operate near the knee of the tradeoff surface: preserving most of the balance gain while avoiding unnecessary movement and continuity disruption.

\section{Evaluation}

We evaluate the framework using two complementary views: historical backtesting and production plan-quality outcomes. The historical backtest assesses whether the core optimization engine can improve territory-plan quality under comparable planning constraints. The production comparison assesses whether adopted plans improved after the decision system was embedded into the latest planning workflow. Across both views, we focus on plan-quality metrics rather than solver objective value alone, because a useful territory plan must be feasible, balanced, interpretable, and operationally usable.

For the historical backtest, we use the previously deployed planning snapshot as the expert-carved baseline and rerun the optimizer using only planning-time information, including the same assignment universe, eligibility rules, locked-account constraints, and capacity bounds available during the original planning cycle. This isolates the plan-quality impact of optimizer-generated scenarios under comparable operating conditions. The adaptive re-optimization workflow described in Section~7 is not simulated in this backtest; it is assessed through the observed production planning outcome.

We assess optimizer output using a plan-quality scorecard rather than solver objective value alone. The scorecard captures the external business properties required for a useful territory design: compliance with planning rules, capture of high-value accounts, opportunity balance across sellers, preservation of valuable continuity, limited account movement, and solver stability within planning runtime constraints. Table~\ref{tab:optimizer_output_evaluation} summarizes these evaluation dimensions.

\begin{table}[t] 
\centering 
\caption{Optimizer output evaluation dimensions.} \label{tab:optimizer_output_evaluation} 
\begin{tabular}{p{0.30\linewidth} p{0.62\linewidth}} 
\toprule 
\textbf{Evaluation area} & \textbf{What it assesses} \\
\midrule 
Compliance & Adherence to eligibility, locked assignments, capacity limits, and business rules. \\ 
Account value & Capture of higher-quality accounts in recommended territories. \\ Balance & Distribution of opportunity across sellers. \\
Continuity & Preservation of valuable account-seller relationships. \\
Movement & Level of account reassignment and operational disruption. \\
Solver stability & Reliability of solutions within planning runtime constraints. \\
\bottomrule 
\end{tabular} 
\end{table}

\begin{figure}[tbh]
  \centering
  \includegraphics[width=\columnwidth]{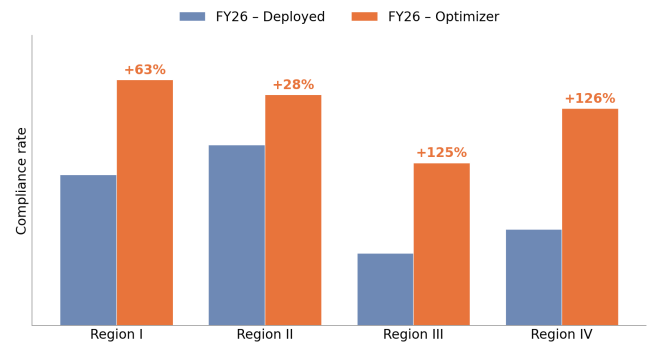}
  \caption{
  Backtest comparison of optimizer-generated and historical territory plans on compliance rate.
  }
  \label{fig:backtest_compliance}
\end{figure}

\begin{figure}[tbh]
  \centering
  \includegraphics[width=\columnwidth]{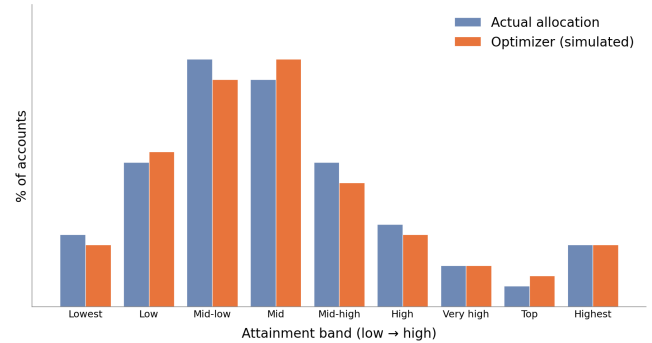}
  \caption{
  Comparison of seller attainment distribution under optimizer-generated and historical territory plans.
  }
  \label{fig:backtest_attainment}
\end{figure}

The optimizer improves assignment compliance by 78\% (Figure~\ref{fig:backtest_compliance}) and produces a modestly more balanced opportunity distribution while preserving comparable seller attainment, with mean attainment essentially unchanged (Figure~\ref{fig:backtest_attainment}). These results indicate that the optimizer improves the feasibility and balance of opportunities without materially distorting the expected seller opportunity under a four-hour solver runtime.

The backtest also estimates the opportunity-value channel from account prioritization. Using the same historical planning snapshot, we compare the account value captured by the optimized plan against the deployed expert-carved plan under comparable coverage capacity. By directing scarce seller coverage toward accounts with higher account values, the optimized plan captures a measurably higher share of total account value than the deployed expert-carved plan under comparable coverage capacity. We interpret this as a directional backtest-based opportunity estimate rather than as realized revenue lift or a randomized causal effect.

We complement the quantitative evaluation with targeted human-in-the-loop review. For selected patches and districts, planning stakeholders inspect recommended assignments, territory distributions and continuity breaks. This review is not a substitute for quantitative evaluation, but a validation step: it helps confirm that the optimizer output is interpretable, actionable, and consistent with planning context that may not be fully captured in the data.

Beyond the historical backtest, the most recent production carve provides an observed deployment outcome from the adopted planning cycle. Relative to the prior deployed plan, the adopted plan improved assignment compliance by 17\% and opportunity balance by 7\%. This production gain is smaller than the backtest improvement by design: the backtest reruns the optimizer cleanly under fixed planning constraints, whereas the production figure reflects the adopted plan after field review, in which expert edits trade some measured compliance for local business judgment not captured in the optimizer. This comparison reflects before-versus-after plan-quality changes in the deployed planning process, not a randomized causal estimate and not a simulated backtest. We use it as evidence that the decision system was operationally adopted and associated with improved plan-quality metrics in production.

\begin{figure*}[t]
  \centering
  \includegraphics[width=\textwidth]{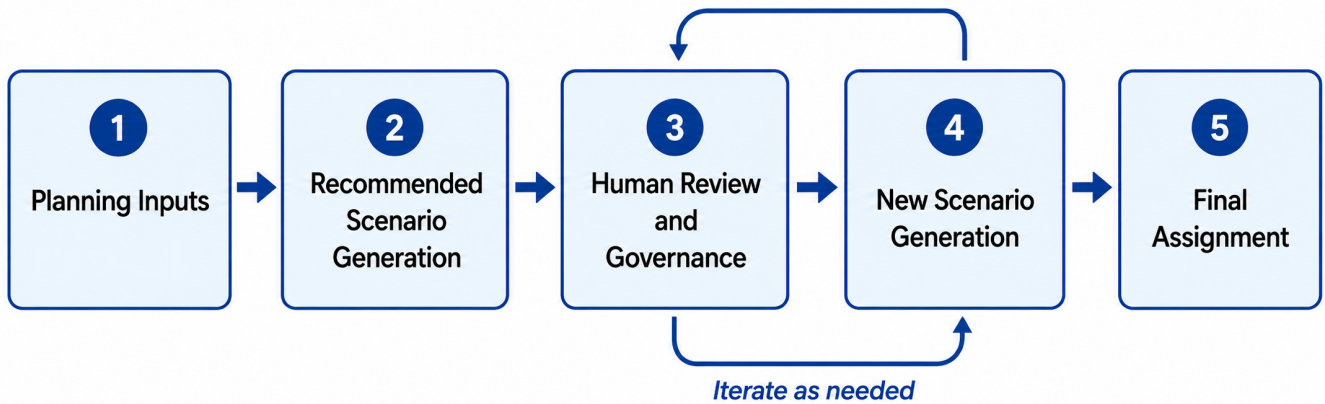}
  \caption{Adaptive human-in-the-loop deployment workflow. Planning inputs drive an initial scenario generation; planners review and govern the recommendation, and re-optimization incorporates their edits, iterating between human review and the optimizer until a final territory plan is adopted.}
  \label{fig:deployment}
\end{figure*}

\section{Deployment}

The framework is operationalized within territory planning as an adaptive, human-in-the-loop decision-support system rather than an automated assignment engine (Figure~\ref{fig:deployment}). At the start of a
planning cycle, the account-value and continuity-valuation layers produce their inputs, which are combined with assignment eligibility and capacity constraints to generate an initial feasible territory
scenario.

As planning progresses, field stakeholders review the recommended scenario and apply expert adjustments that reflect local context and business priorities not fully captured in the data. The optimizer is
periodically re-run against the latest reviewed plan, so that algorithmic recommendations and accumulated human edits continuously inform one another; the intent-to-keep term (Section~\ref{sec:optimization})
prevents these refreshes from reverting deliberate expert changes.

Optimizer outputs are surfaced as inspectable planning scenarios rather than written directly as final assignments, allowing stakeholders to compare a recommendation against the current plan and decide
whether to adopt, adjust, or use it as reference. The account-scoring layer is additionally used on its own for account-value inspection and territory review. Together, this shifts territory carving from
account-by-account manual assignment toward scenario-based governance that embeds predictive scoring, continuity valuation, and constrained optimization into a governable decision system---improving
consistency and scale while preserving human control.

\section{Lessons Learned}
First, objective-function design is the core challenge. Invoking an optimizer is not sufficient; the difficult work is translating business priorities into measurable objective terms and constraints. Account value, continuity, balance, and capacity are not naturally expressed in the same units, so the objective must be carefully calibrated to avoid over-weighting one dimension.

Second, not every business rule should be a hard constraint. Some requirements, such as segment eligibility or locked assignments, may need to be enforced strictly. Others, such as balance targets or continuity preferences, may be better represented as soft penalties that allow the optimizer to make tradeoffs when necessary.

Third, continuity should be empirically calibrated where possible. Planning teams often know that relationship disruption matters, but the magnitude is difficult to reason about. Synthetic-control-based valuation helps convert this intuition into a measurable tradeoff that can be used in optimization.

Fourth, common currency matters for adoption. When account scores, continuity values, and balance penalties are expressed only in abstract optimization units, stakeholders may find it difficult to understand why the optimizer prefers one assignment over another.

Finally, human-in-the-loop design is essential because territory carving involves strategic context that may not be fully captured in data. The optimizer should therefore serve as a consistent baseline for expert review, not as an automatic final decision.

\section{Limitations}
\label{sec:limitations}

Several limitations bound the claims in this paper.

\paragraph{Account value is treated as assignment-invariant.}
The optimizer maximizes captured account value under the assumption that an account's value is independent of which seller is assigned to it. In practice, seller fit, capacity, and continuity all plausibly affect realized outcomes; relaxing this assumption---jointly modeling account value and assignment---is an open problem.

\paragraph{Continuity is estimated, not identified.}
As discussed in Section~\ref{sec:continuity}, the handoff effect is observational. The placebo adjustment guards against common time-period effects but not against selection into handoff on unobservables or cross-account spillovers. The reported ATT should be read as directional.

\paragraph{Evidence type.}
Our evidence combines historical backtesting and production before-versus-after comparisons. The backtest results are backtest-based comparisons to the deployed expert plan, while the production compliance and balance gains are observed outcomes from adopted production plans. The production results demonstrate operational improvement but should not be read as randomized causal estimates, since planning rules, account mix, and field behavior may differ across cycles. The account-value opportunity estimate is also directional rather than measured realized lift.

\paragraph{Operational scope.}
The current formulation does not encode patch-level rules such as geography or language coverage, can produce weaker plans in low density districts where few feasible sellers exist, and may surface infeasibility when a district mixes segments or sales motions. These cases presently fall back to human review.

\paragraph{Toward causal validation.}
The natural next step is to move from observational backtesting to staged or randomized rollout---varying assignment policy across comparable districts---so that compliance, balance, and continuity effects can be estimated causally rather than inferred from a single planning snapshot.

\section{Conclusion}
We present a production-oriented decision system for territory carving that combines account value quantification, continuity valuation, and multi-objective optimization. The system has been operationalized in planning workflows as a scenario-based decision-support tool: optimizer outputs serve as recommended planning scenarios that stakeholders can adopt, adjust, or use as references, while account scores also support independent account-value review. The broader contribution is less any single component than the reframe they enable: by expressing account value as a common currency and continuity as a measured cost, territory carving becomes a transparent, governable decision surface rather than a distributed act of expert judgment.

Future work will focus on further improving the multi-objective formulation, including additional realistic planning constraints, more interpretable objective terms, and stronger explanation of optimizer recommendations. We also plan to extend optimization from account-to-seller assignment toward higher-level district and regional planning, where larger planning scopes make manual tradeoff management increasingly difficult.

\begin{acks}
We are especially grateful to the Strategic Planning (SPACE) team for their close partnership throughout this work. Their domain expertise, planning-design guidance, and support in driving stakeholder adoption were central to translating the decision framework into a usable territory-planning system. We thank Megan Luu, Andrew Liu, Hannah Wilenskey-Ho, Marcus Trott, Stephen Huang, Sid Kumar, Sahana Sarma, and the broader SPACE team for their sustained collaboration.

We also thank Jake Brutlag, Jason Huang, Feng Pan, Divy Menghani and our internal data science reviewers for valuable feedback on the framing, methodology, and evaluation. We thank Vinson Wei for analytical and implementation support. We are grateful to the GTM Analytics team, including Petar Zoric and David Gojo, for related input on the account-signal foundation. Finally, we thank the territory-planning stakeholders whose domain review helped validate the system and shape its production use.
\end{acks}

\bibliographystyle{ACM-Reference-Format}
\bibliography{references}


\begin{thebibliography}{9}


\ifx \showCODEN    \undefined \def \showCODEN     #1{\unskip}     \fi
\ifx \showISBNx    \undefined \def \showISBNx     #1{\unskip}     \fi
\ifx \showISBNxiii \undefined \def \showISBNxiii  #1{\unskip}     \fi
\ifx \showISSN     \undefined \def \showISSN      #1{\unskip}     \fi
\ifx \showLCCN     \undefined \def \showLCCN      #1{\unskip}     \fi
\ifx \shownote     \undefined \def \shownote      #1{#1}          \fi
\ifx \showarticletitle \undefined \def \showarticletitle #1{#1}   \fi
\ifx \showURL      \undefined \def \showURL       {\relax}        \fi
\providecommand\bibfield[2]{#2}
\providecommand\bibinfo[2]{#2}
\providecommand\natexlab[1]{#1}
\providecommand\showeprint[2][]{arXiv:#2}

\bibitem[Abadie(2021)]%
        {abadie2021using}
\bibfield{author}{\bibinfo{person}{Alberto Abadie}.}
  \bibinfo{year}{2021}\natexlab{}.
\newblock \showarticletitle{Using Synthetic Controls: Feasibility, Data
  Requirements, and Methodological Aspects}.
\newblock \bibinfo{journal}{\emph{Journal of Economic Literature}}
  \bibinfo{volume}{59}, \bibinfo{number}{2} (\bibinfo{year}{2021}),
  \bibinfo{pages}{391--425}.
\newblock


\bibitem[Abadie et~al\mbox{.}(2010)]%
        {abadie2010synthetic}
\bibfield{author}{\bibinfo{person}{Alberto Abadie}, \bibinfo{person}{Alexis
  Diamond}, {and} \bibinfo{person}{Jens Hainmueller}.}
  \bibinfo{year}{2010}\natexlab{}.
\newblock \showarticletitle{Synthetic Control Methods for Comparative Case
  Studies: Estimating the Effect of California's Tobacco Control Program}.
\newblock \bibinfo{journal}{\emph{J. Amer. Statist. Assoc.}}
  \bibinfo{volume}{105}, \bibinfo{number}{490} (\bibinfo{year}{2010}),
  \bibinfo{pages}{493--505}.
\newblock


\bibitem[Gupta et~al\mbox{.}(2006)]%
        {gupta2006}
\bibfield{author}{\bibinfo{person}{Sunil Gupta}, \bibinfo{person}{Dominique
  Hanssens}, \bibinfo{person}{Bruce Hardie}, \bibinfo{person}{William Kahn},
  \bibinfo{person}{V. Kumar}, \bibinfo{person}{Nathaniel Lin},
  \bibinfo{person}{Nalini Ravishanker}, {and} \bibinfo{person}{S. Sriram}.}
  \bibinfo{year}{2006}\natexlab{}.
\newblock \showarticletitle{Modeling Customer Lifetime Value}.
\newblock \bibinfo{journal}{\emph{Journal of Service Research}}
  \bibinfo{volume}{9}, \bibinfo{number}{2} (\bibinfo{year}{2006}),
  \bibinfo{pages}{139--155}.
\newblock


\bibitem[Hervert-Escobar and Alexandrov(2018)]%
        {hervert2018}
\bibfield{author}{\bibinfo{person}{Laura Hervert-Escobar} {and}
  \bibinfo{person}{Vassil Alexandrov}.} \bibinfo{year}{2018}\natexlab{}.
\newblock \showarticletitle{Territorial Design Optimization for Business Sales
  Plan}.
\newblock \bibinfo{journal}{\emph{J. Comput. Appl. Math.}}
  \bibinfo{volume}{340} (\bibinfo{year}{2018}), \bibinfo{pages}{501--507}.
\newblock


\bibitem[Jena et~al\mbox{.}(2023)]%
        {jena2023}
\bibfield{author}{\bibinfo{person}{Suvendu Jena}, \bibinfo{person}{Jilei Yang},
  {and} \bibinfo{person}{Fangfang Tan}.} \bibinfo{year}{2023}\natexlab{}.
\newblock \showarticletitle{Unlocking Sales Growth: Account Prioritization
  Engine with Explainable AI}.
\newblock \bibinfo{journal}{\emph{arXiv preprint arXiv:2306.07464}}
  (\bibinfo{year}{2023}).
\newblock


\bibitem[Miettinen(1999)]%
        {miettinen1999}
\bibfield{author}{\bibinfo{person}{Kaisa Miettinen}.}
  \bibinfo{year}{1999}\natexlab{}.
\newblock \bibinfo{booktitle}{\emph{Nonlinear Multiobjective Optimization}}.
\newblock \bibinfo{publisher}{Kluwer Academic Publishers}.
\newblock


\bibitem[R{\'i}os-Mercado and L{\'o}pez-P{\'e}rez(2013)]%
        {riosmercado2013}
\bibfield{author}{\bibinfo{person}{Roger~Z. R{\'i}os-Mercado} {and}
  \bibinfo{person}{J.~Fabi{\'a}n L{\'o}pez-P{\'e}rez}.}
  \bibinfo{year}{2013}\natexlab{}.
\newblock \showarticletitle{Commercial Territory Design Planning with
  Realignment and Disjoint Assignment Requirements}.
\newblock \bibinfo{journal}{\emph{Omega}} \bibinfo{volume}{41},
  \bibinfo{number}{3} (\bibinfo{year}{2013}), \bibinfo{pages}{525--535}.
\newblock


\bibitem[Skiera and Albers(1998)]%
        {skiera1998}
\bibfield{author}{\bibinfo{person}{Bernd Skiera} {and}
  \bibinfo{person}{S{\"o}nke Albers}.} \bibinfo{year}{1998}\natexlab{}.
\newblock \showarticletitle{{COSTA}: Contribution Optimizing Sales Territory
  Alignment}.
\newblock \bibinfo{journal}{\emph{Marketing Science}} \bibinfo{volume}{17},
  \bibinfo{number}{3} (\bibinfo{year}{1998}), \bibinfo{pages}{196--213}.
\newblock


\bibitem[Zoltners and Sinha(1980)]%
        {zoltners1980}
\bibfield{author}{\bibinfo{person}{Andris~A. Zoltners} {and}
  \bibinfo{person}{Prabhakant Sinha}.} \bibinfo{year}{1980}\natexlab{}.
\newblock \showarticletitle{Integer Programming Models for Sales Resource
  Allocation}.
\newblock \bibinfo{journal}{\emph{Management Science}} \bibinfo{volume}{26},
  \bibinfo{number}{3} (\bibinfo{year}{1980}), \bibinfo{pages}{242--260}.
\newblock


\end{thebibliography}

\end{document}